\documentclass[fleqn,10pt]{wlscirep}
\usepackage[utf8]{inputenc}
\usepackage[T1]{fontenc}
\usepackage{textgreek}
\usepackage{lineno}
\usepackage{amsmath}

\title{Convolution Neural Network based Mode Decomposition for Degenerated Modes via Multiple Images from Polarizers}

\author[1,*]{Hyuntai Kim}
\affil[1]{Department of Electronic and Electrical Converged Engineering, Hongik University, Sejong 30016, Korea}

\affil[*]{corresponding author: Hyuntai Kim (hyuntai@hongik.ac.kr)}

\begin{abstract}
In this paper, a mode decomposition (MD) method for degenerated modes has been studied. Convolution neural network (CNN) has been applied for image training and predicting the mode coefficients. Four-fold degenerated $LP_{11}$ series has been the target to be decomposed. Multiple images are regarded as an input to decompose the degenerate modes. Total of seven different images, including the full original near-field image, and images after linear polarizers of four directions (0$^\circ$, 45$^\circ$, 90$^\circ$, and 135$^\circ$), and images after two circular polarizers (right-handed and left-handed) has been considered for training, validation, and test. The output label of the model has been chosen as the real and imaginary components of the mode coefficient, and the loss function has been selected to be the root-mean-square (RMS) of the labels. The RMS and mean-absolute-error (MAE) of the label, intensity, phase, and field correlation between the actual and predicted values have been selected to be the metrics to evaluate the CNN model. The CNN model has been trained with 100,000 three-dimensional images with depths of three, four, and seven. The performance of the trained model was evaluated via 10,000 test samples with four sets of images - images after three linear polarizers (0$^\circ$, 45$^\circ$, 90$^\circ$) and image after right-handed circular polarizer - showed 0.0634 of label RMS, 0.0292 of intensity RMS, 0.1867 rad of phase MAE, and 0.9978 of average field correlation. The performance of 4 image sets showed at least 50.68\% of performance enhancement compared to models considering only images after linear polarizers.
\end{abstract}

\begin{document}

\flushbottom
\maketitle

\thispagestyle{empty}

\section{Introduction}
Because of the boost in data traffic and intensity of high-power lasers, the demands for multimode optical waveguides have been increasing \cite{fob,mmf-comm,mmf-hp}. multimode waveguides have an advantage not only in communications and high-power transmissions, but also in various application fields such as imaging, nonlinear optics, and quantum physics \cite{mmf-im,mmf-nle,mmf-qm}. In specific, multimode waveguide is capable to generate or transmit radially polarized light, which is suitable for optical trapping, machining, and longitudinal focusing \cite{rpg1,rpg2,rpg3,sol1,sol2,rps1,rps2}. To understand the ratio of different modes, modal decomposition (MD) technique is inevitable. MD has been studied intensively, and achieved in various methods, such as cross-correlated imaging, spatially and spectrally resolved imaging, low-coherence interferometry, multi-variable optimization algorithm, and correlation filter \cite{md-c2,md-ss,md-lci,md-mvoa, md-cgf}.
\\Thanks to various algorithms and increase in computing power, artificial intelligence has been explosively studied and applied in various fields, including the optics research field \cite{aip1,aip2}. Especially, convolution neural network (CNN) technique which is proper for image process has been applied in various fields, including MD\cite{cnn-md1,cnn-md2,cnn-md3}. [\citenum{cnn-md1}] has studied MD for one polarization state from near-field image, and [\citenum{cnn-md2}] has applied far-field images as additional input data. However, these existing studies have not considered degenerated modes. $LP_{01}$ mode is two-fold degenerated, and $LP_{11}$ mode is four-fold degenerated, which are $TM_{01}, TE_{01}$, and two $HE_{21}$ modes. To obtain the specific mode ratio such as radially polarized mode, the coefficient of each mode is required. These degenerated modes show exactly identical near-field profiles, therefore it is difficult to exactly decompose the degenerated modes by a single near-field image. 
\\Recently, [\citenum{cnn-md3}] has studied MD for degenerated modes using polarization filters. The results were remarkable which showed 99.1\% of field correlations. The study used three polarized images, and the $LP_{11}$ mode is a four-fold degenerated state, therefore three polarized images are the minimum input to examine the weights of four modes. However, as the image shows only the intensity of the complex field, images after LPs are unable to distinguish the rotating direction of circular or elliptical polarized lights.
\\In this paper, images after linear polarizers (LPs) and circular polarizers (CPs) are applied to compare the degenerated modes. Four $LP_{11}$ groups are considered, and six additional near-field images after LPs and CPs are considered.  Four images after LPs of direction with $x \:(0^\circ)$, $x+y\: (45^\circ)$, $y\: (90^\circ)$, and $x-y \:(135^\circ)$, and two images after right-handed CP (RHCP) and left-handed CP (LHCP) are considered. Among the seven images, three, four, and seven sets of images are chosen to train the CNN model. Eigenmodes have been obtained and reproduced based on numerical simulations, and sample images are also calculated based on eigenmodes from numerical results. The images are stacked to form a three-dimensional tensor with a depth of three, four, and seven and are regarded as an input of CNN, and the model is trained by generated images.

\section{Principles and Methods}
\subsection{Data preparation}
The electric field of waveguide can be decomposed by linear combination of its eigenmodes, such as following equation.
\begin{equation}
\vec{E}_{net}(r,\theta)= \sum_{n=1}^{N} C_n \vec{E}_n(r,\theta) ,
\end{equation}
where $C_n$ is the complex coefficient of the eigenmode $\vec{E}_n$. The complex coefficient $C_n$ can be expressed as both intensity-phase form and real-imaginary form, such as following equation.

\begin{equation}
C_n=\rho_n exp(j \phi_n) = x_n + j y_n
\end{equation}

\begin{figure}[!htb]
\centering
\includegraphics[width=0.9\textwidth]{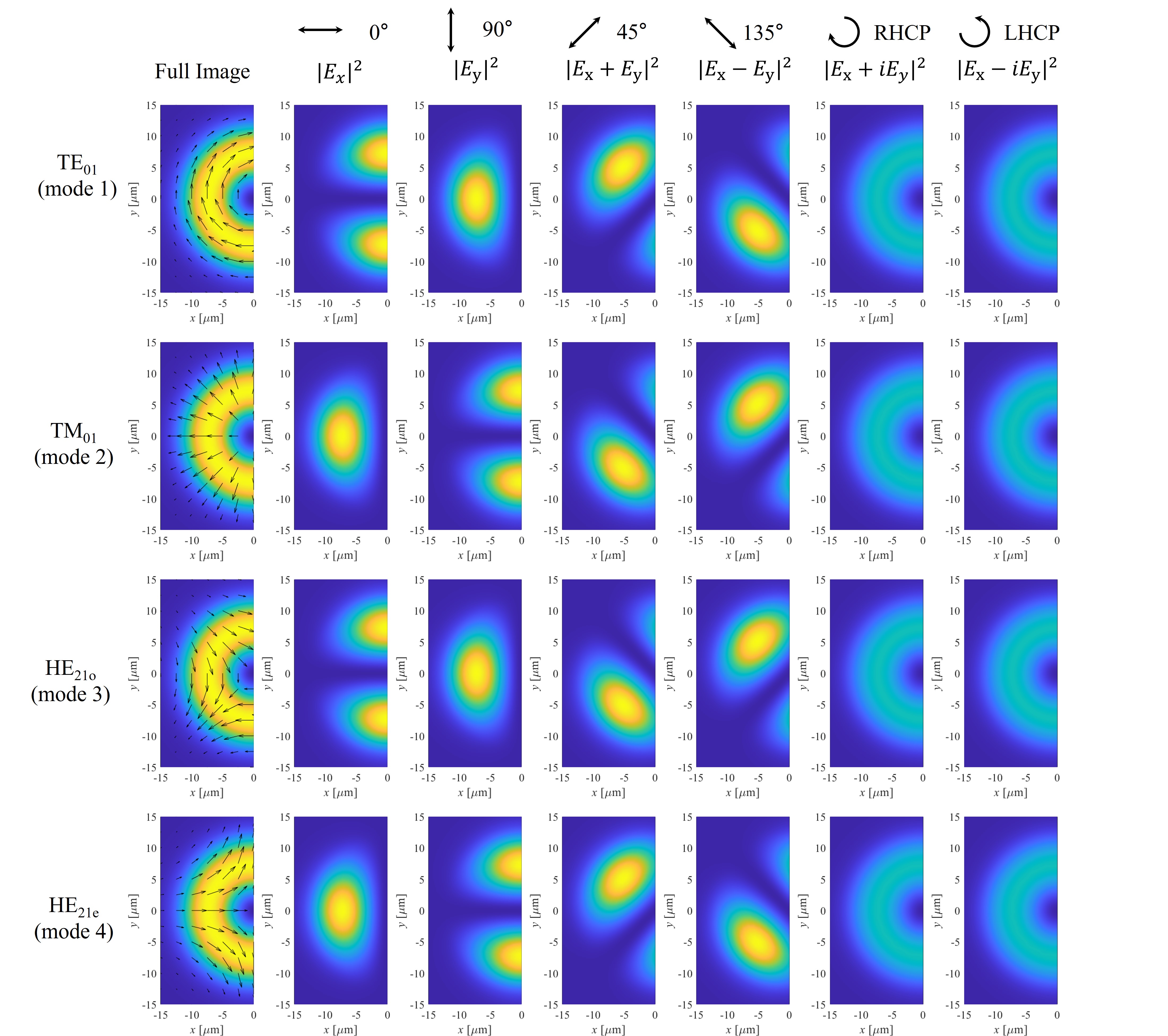}
\caption{Seven images of $LP_{11}$ series eigenmodes. The electric vector field is depicted within the full image.}
\label{fig:1}
\end{figure}

The modes of interest in this paper are $LP_{11}$ series, which are $TE_{01}, TM_{01}$, and two $HE_{21}$ modes. Each mode can be expressed as the following equations.
 
\begin{align}
  \begin{split}
&\vec{E}_{TE_{01}}(r,\theta)=B(r)(-sin\theta a_x+cos\theta a_y)
\\
&\vec{E}_{TM_{01}}(r,\theta)=B(r)(cos\theta a_x+sin\theta a_y)
\\
&\vec{E}_{HE_{21o}}(r,\theta)=B(r)(-sin\theta a_x-cos\theta a_y)
\\
&\vec{E}_{HE_{21e}}(r,\theta)=B(r)(-cos\theta a_x+sin\theta a_y)
  \end{split}
\end{align}
 where $B(r)$ is the radial distribution of $LP_{11}$, which is generally a Bessel function \cite{lpm}.
The mode profile of $LP_{11}$ series has been calculated via finite element method via COMSOL. The multimode waveguide has been selected to be a conventional multimode fiber, FG025LJA (Thorlabs), a step-index multimode fiber with 0.1 of NA, and a core diameter of 25 um. The operation wavelength has been assumed to be 1064 nm, which is a typical value for Yb doped lasers \cite{ydf}.
Hereinafter, the four modes, $TE_{01}, TM_{01}, HE_{21o}$, and $HE_{21e}$ are indexed as mode 1, 2, 3, and 4, respectively. The degenerated $LP_{11}$ series can be expressed by four imaginary coefficients $C_1, C_2, C_3$, and $C_4$. The net intensity $|\vec{E}^2|$ is not phase-sensitive, so one could tune one coefficient to be real, or phase to be 0. Hereinafter, the phase is tuned as $C_1$ to be a positive real number, which indicates $\phi_1=0$, $x_n>0$, and $y_1=0$.
Now to generate a degenerated sample, four random coefficients are generated. The coefficients are weighed to each eigenmode and the final near-field image is generated. Not only the image without the polarizer, but images after four LPs – $x \:(0^\circ)$, $x+y\: (45^\circ)$, $y\: (90^\circ)$, and $x-y \:(135^\circ)$ – and two CPs with different rotation directions - RHCP and LHCP - are also calculated and generated. The pixel number of the generated image has been chosen to be 121$\times$121, where the core region has been divided into 100 pieces, and 10\% margin on both sides was regarded as the area of interest. However, as discussed previously, the left half of the image has an identical tendency to the right half, so taking half of the image is enough for comparison on degenerated $LP_{11}$ series MD. Therefore, the tenor size of the final input for training becomes 61$\times$121$\times n$, where $n$ is the number of images selected in a single case. Generated seven images for four eigenmodes are shown in Fig. \ref{fig:1}. The electric vector field is shown within the unpolarized full image.

\begin{figure}[!htb]
\centering
\includegraphics[width=0.9\textwidth]{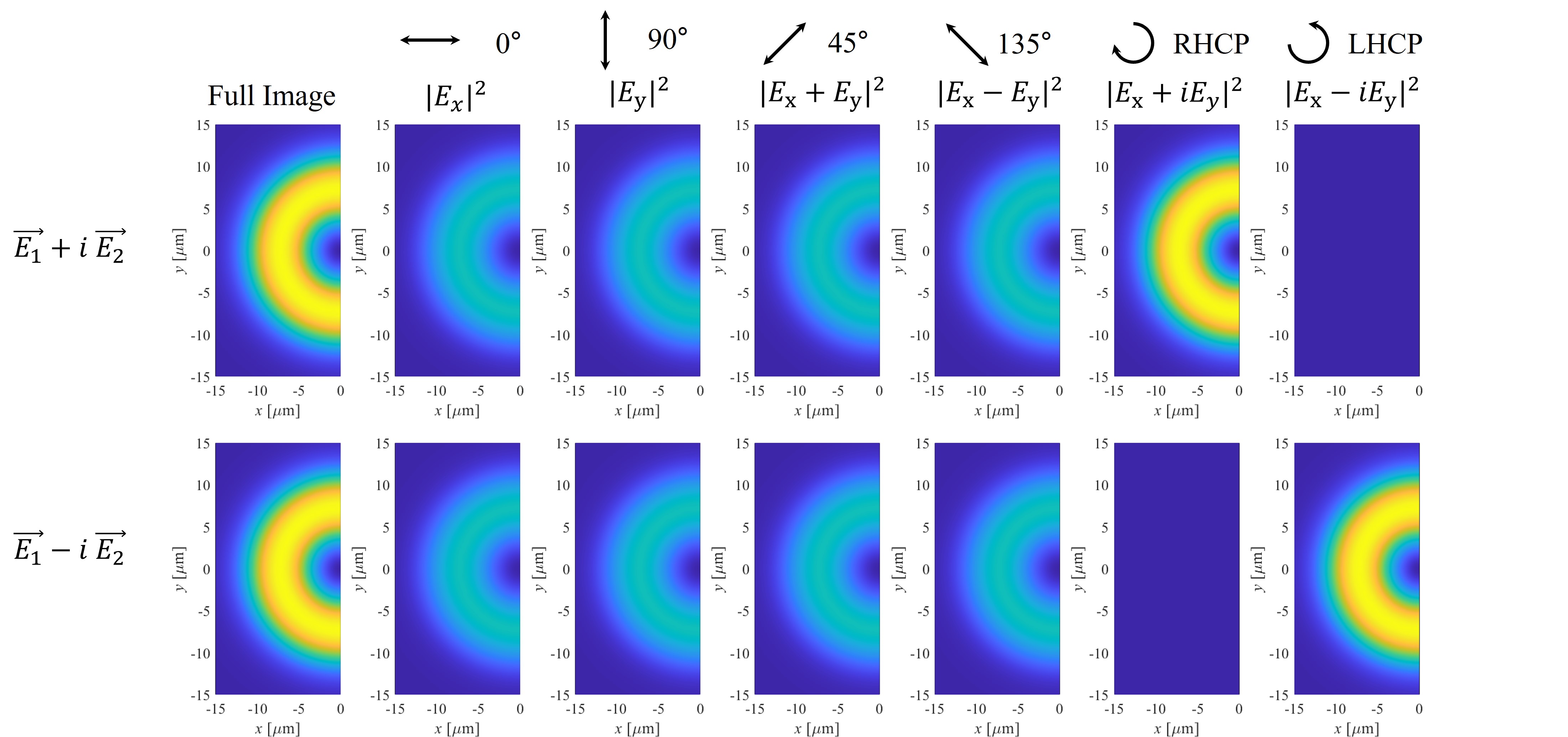}
\caption{Seven images of twp mixed $LP_{11}$ series, $\vec{E_1}+i\vec{E_2}$ and $\vec{E_1}-i\vec{E_2}$.}
\label{fig:1a}
\end{figure}

From Fig. \ref{fig:1}, it seems that the modes are available to be compared only by using three images after three LPs. As the degenerated mode has four unknown coefficients $C_1 \sim C_4$, and the problem is to find the ratio between four variables, three equations are enough to solve the problem. However, the image only shows the absolute value not the exact complex value, three images are not enough for some case. Fig. \ref{fig:1a} shows two different mixed modes.
 The first and second mixed mode is chosen to be sum of $TE_{01}$ mode and $TM_{01}$ mode with different phase constant, which can be expressed as $\vec{E_1}-i\vec{E_2}$, respectively. As the electric field direction of $TE_{01}$ mode and $TM_{01}$ mode are perpendicular in all the regions, the electric field of mixed mode becomes circularly polarized in every positions. Therefore, it is unavailable to distinguish two modes only via LP. In other words, the MD system using LP cannot compare two perpendicular modes have leading phase or retarding phase. It is shown in Fig. \ref{fig:1a} that two different modes show exactly the same images after all directions of LPs. In this paper, four sets of images and seven sets of images are considered to be the input. Four images are selected to be the images after LPs of angle with $x \:(0^\circ)$, $x+y\: (45^\circ)$ and $y\: (90^\circ)$, and an additional image after RHCP. Seven sets are chosen to be all the images depicted in Fig. \ref{fig:1} and Fig. \ref{fig:1a}. For comparison, case of three LPs are also tested.

\subsection{CNN Model}
\begin{figure}[!htb]
\centering
\includegraphics[width=0.9\textwidth]{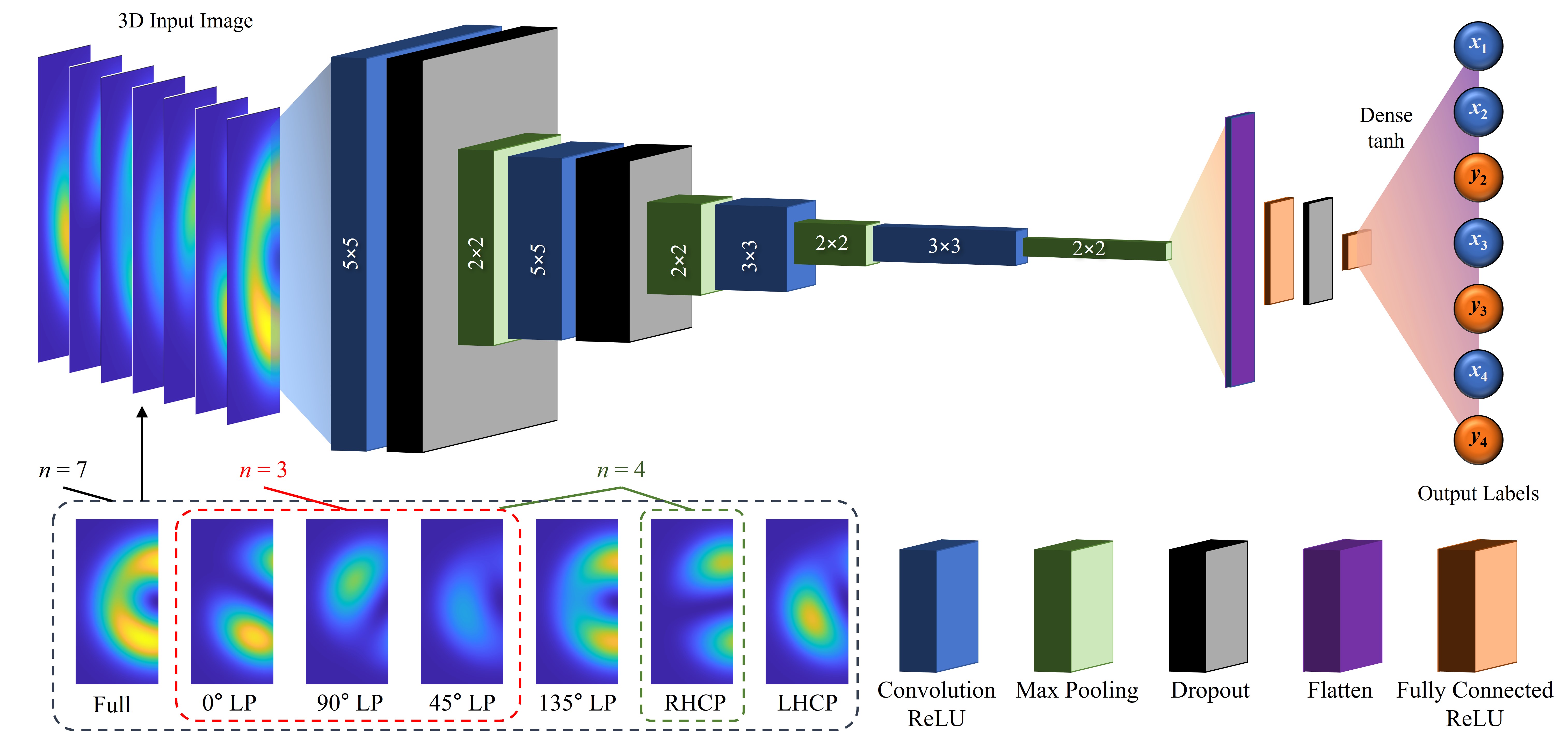}
\caption{The schematic of the full CNN system. 3D tensors with depth of three, four, and seven are inserted into the CNN architecture and 7 labels, the real and image value of each mode coefficients, are obtained as an output.}
\label{fig:2}
\end{figure}

The CNN model has been constructed for MD. The size of the input is selected to be three, four, and seven groups of 61$\times$121 images, resulting in 61$\times$121$\times n$ images where $n=$3,4, and 7.  
The input image goes through a CNN network, which is composed of four convolution rectified linear unit (ReLU) layers, three max pooling layers, a flatten layer, two fully connected ReLu layers, three dropout layers, and the output target of 7 tanh layer. To avoid overfitting, all layers have $10^{-8}$ L2 regulation, and the dropout ratio of 5\% has been applied. The full system is shown in Fig. \ref{fig:2}. Note that the numbers at the convolution block of Fig. \ref{fig:2} represent the filter size of the convolution tensor, and the size of the block represents the shape of the output after the layer.
 The loss of the model was selected to be root-mean-square (RMS), or mean-squared-error (MSE). In terms of loss function, [\citenum{cnn-md2}] has used field correlation function. However, the degenerated case shows similar correlation functions, so in this paper, the loss function is assumed to be RMS between the values. It is notable that the phase and intensity are different units, so if coefficient $\rho$ and $\phi$ are used to measure loss such as [\citenum{cnn-md1}], one must carefully determine the weight of intensity and phase. It is reasonable to assume the loss between the actual coefficient $C_{n,a}$ and the predicted coefficient $C_{n,p}$ to be the distance between the two imaginary values. The distance is available to be expressed as $\sqrt{(x_{n,a}-x_{n,p})^2+(y_{n,a}-y_{n,p})^2}$, so if the output label is determined to be $x_1, x_2, y_2, x_3, y_3, x_4$, and $y_4$, the RMS or MSE will represent the distance between the answer and predicted value. The labels are normalized, the largest value to become 1. Labels are denoted as $z_n$, where $z_{1,2,4,6}=x_{1,2,3,4}$ and $y_{2,3,4}=z_{3,5,7}$. Note that the first mode is tuned to be real, so $y_1$ is always 0, which can be removed from the output label. The RMS loss of real and complex components, will be referred to hereinafter as label RMS. The label RMS is shown in the following equation.
\begin{equation}
    Label \:RMS = \sqrt{\frac{\sum_{n=1}^{4} (x_{n,a}^2-x_{n,p}^2) + \sum_{n=2}^{4} (y_{n,a}^2-y_{n,p}^2) }{7}}=\sqrt{\overline{\sum_{n=1}^{7} (z_{n,a}^2-z_{n,p}^2) }}=\sqrt{\overline{\Delta z^2}}
\end{equation} 
The optimizer of model has been chosen to be “ADAM”, and the batch size was selected as 128. 300 epoch of training was performed with 100,000 training sets and 3,000 validation sets.
After training, the mode intensity ratio $\rho$, phase $\phi$ and reconstructed field images have been calculated from the predicted labels. The metric of the model was evaluated in terms of RMS, mean-absolute-error (MAE) and the field correlation between the actual and seven reconstructed near-field images, which are the original full image and six images after the polarizers.

\section{Results}

\begin{figure}[!htb]
\centering
\includegraphics[width=0.55\textwidth]{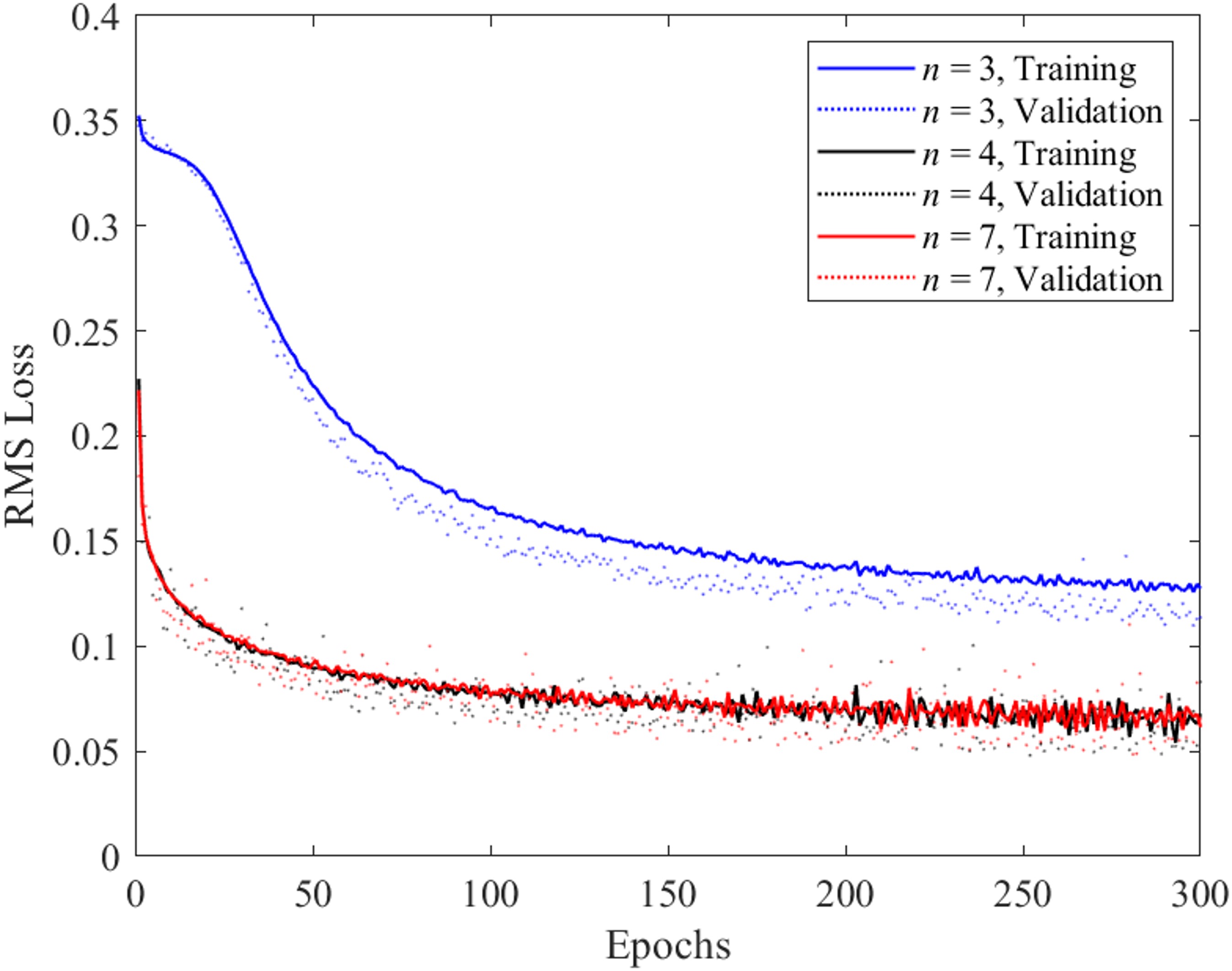}
\caption{ The RMS loss history while training, in terms of epochs for case of $n=$3, 4, and 7. Both RMS losses of the training set and validation set are depicted. }
\label{fig:3}
\end{figure}

The model is trained with 100,000 generated 61$\times$121$\times n$ training images with 3000 validation sets, and the RMS loss in terms of epoch is shown in Fig. \ref{fig:3} for three cases, $n=$3, 4, and 7. RMS of the training set and validation set are depicted, respectively. It is clear that the case of $n=3$, which only considers images after LPs, converges slower, and shows higher loss compared to the cases which consider image after CP. Both case of $n=$4 and 7, which consider images after CP, shows similar converge speed and losses.
 
To evaluate the performance of the model, the model predicted 10,000 new test sets. From the predicted seven labels, obtain four weight coefficients ($\rho_1 \sim \rho_4$) and three relative phases($\phi_2 \sim \phi_4$), and the field correlation of seven images are inversely calculated. The label MAE ($\overline{|\Delta z|}$), label RMS ($\sqrt{\overline{\Delta z^2}}$), MAE and RMS of weight coefficient ($\overline{|\Delta \rho|}$ and $\sqrt{\overline{\Delta\rho^2}}$), 
normalized average difference (MAE) of phase ($\overline{|\Delta \phi|}/2\pi$), the field correlation of the full unpolarized image ($Corr_{full}$), and the average of seven field correlations ($\overline{Corr}$) have been calculated and shown in Table \ref{t1}. Note the angle difference angle $\Delta\rho$ has been calculated from the angle of $C_{n,a}/C_{n,p}$. It is shown that all the cases of $n=$4 and 7 which considered images after CP show better performance compared to the case of $n=3$ which only considered images after LP. Especially the average correlation of $n=3$ is less than 99\%. Because the case of $n=3$ is unable to compare the direction of circular or Elliptical polarized light, the field correlations of images after RHCP and LHCP are relatively low, which are calculated as 0.9855 and 0.9859. The performance enhancement between the case of $n=$3 and 4 is also calculated and shown in the last row in Table \ref{t1}. Note that the loss of correlation has been regarded as $1-Corr$. It is shown that all the metrics showed at least 50\% of reduced loss, which indicates the case with the CP predicts the degenerated mode with higher accuracy. 

\begin{center}
\begin{tabular}{ c ||c | c| c| c| c| c| c }
   & $\overline{|\Delta z|}$ & $\sqrt{\overline{\Delta z^2}}$ & $\overline{|\Delta \rho|}$ &$\sqrt{\overline{\Delta\rho^2}}$ & $\overline{|\Delta \phi|}/2\pi$ & $Corr_{full}$ & $\overline{Corr}$\\ 
 \hline
 \hline
 $n=3$ & 0.0999&0.0773&0.0516&0.0435&0.0449&0.9949&0.9887\\  
 $n=4$ & 0.0634&0.0513&0.0291&0.0247&0.0297&0.9978&0.9963\\
 $n=7$ & 0.0516&0.0417&0.0236&0.0200&0.0244&0.9988&0.9976 \\
 \hline
$n=3$ vs 4 (\%) &57.57&50.68&77.32&76.11&51.18&131.82&205.41
\end{tabular}
\captionof{table}{Calculated metrics for $n=$3, 4 and 7.}\label{t1}
\end{center}

\begin{figure}[!htb]
\centering
\includegraphics[width=0.95\textwidth]{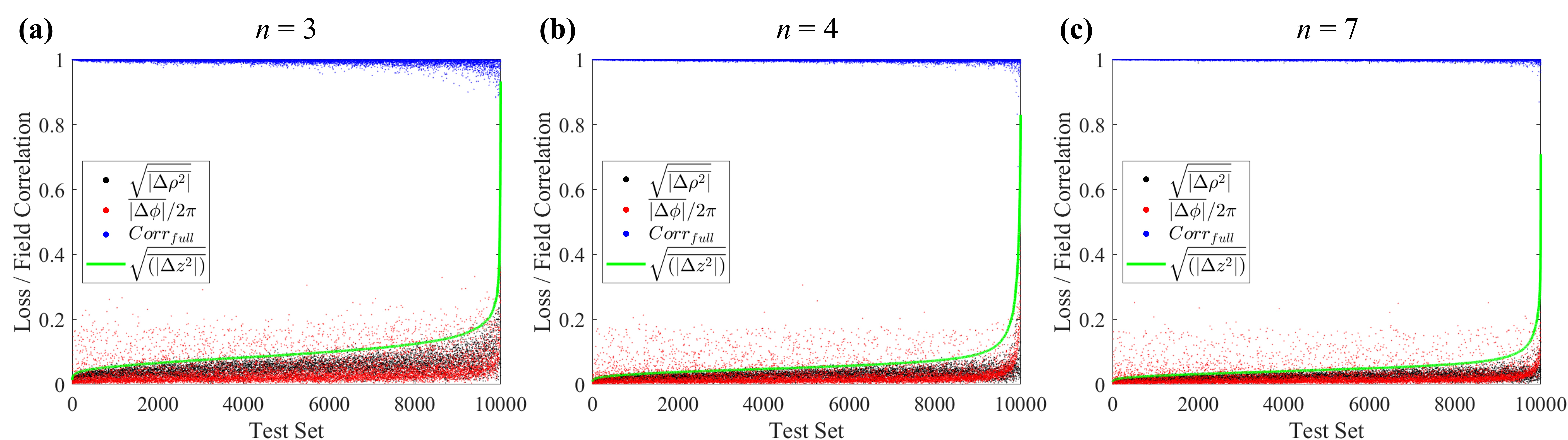}
\caption{ The intensity RMS loss (red dots), normalized phase RMS loss (black dots), and field correlation (blue dots) of 10,000 test sets for (a) $n=3$, (b) $n=4$, and (c) $n=7$. The metrics are sorted by the label RMS loss (green solid line).}
\label{fig:3a}
\end{figure}

The metrics – RMS of weight coefficient ($\sqrt{\overline{\Delta\rho^2}}$), MAE of normalized phase ($\overline{|\Delta \phi|}/2\pi$), and field correlation of the full image ($Corr_{full}$) – sorted by the label RMS ($\sqrt{\overline{\Delta z^2}}$), are depicted in Fig. \ref{fig:3a} for the case of $n=$3, 4, and 7. It is shown that low label loss tends to show high field correlation, and low mode weight MAE and phase MAE. 
However, the metrics are not exactly correlated to the label MAE, where some of the outliers are observed where low label MAE loss has high intensity/phase loss or low field correlation. The fact indicates the field correlation is not the best metric to be optimized while training, where similar mode images can be generated from different combinations. One may use the seven field correlations as a metric, however, it would add complexity to the calculation.
Comparing the three cases, it is obvious that the case of $n=3$ shows the highest loss. The label RMS (green solid line) has a higher value compared to other cases ($n=4,7$). The intensity weight and phase difference of $n=3$ also show more outliers. For the case of $n=4$ and 7, most of intensity and phase losses (red and black dots) are positioned at the bottom line and the field correlation (blue dots) is mostly shown at the top line. From the results, it is clear that the cases which considered the image after CP shows higher performance. The case of $n=$4 and 7 shows similar performance, so the case of four sets of images is assumed to be proper which requires low memory and calculation time.

\begin{figure}[!htb]
\centering
\includegraphics[width=0.95\textwidth]{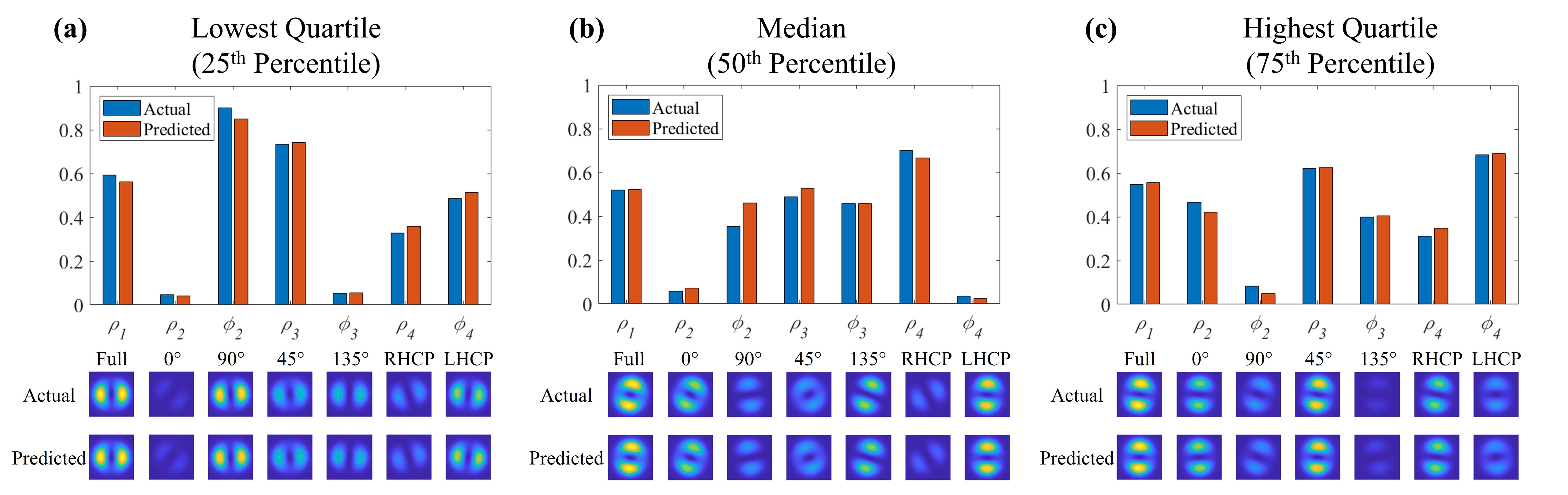}
\caption{Comparison between intensity $\rho$ and phase $\phi$ of mode coefficient and their seven images for (a) lowest quartile, (b) median, and (c) highest quartile.}
\label{fig:4}
\end{figure}

The comparison of modal weight $\rho$, phase $\phi$, and the seven images for the case of $n=4$ are shown in Fig. \ref{fig:4}. The samples were chosen as the lowest quartile, median, and highest quartile in terms of the label MAE loss – which has $2500^{th}, 5000^{th}$, and $7500^{th}$ smallest label MAE loss. It is clear that the coefficients and the images show relatively good matching between the actual value/image and predicted ones. 

\section{Discussions}
The MD study based on CNN has not been thoroughly studied. The metrics, loss function, and method to verify the model are not standardized. For in case of loss function, the results show that field correlation is not a proper metric to understand the exact ratio of each mode. Using both intensity and phase, the weight function on different dimensions is another hyper-parameter to be optimized. Therefore, the paper proposes the real and imaginary component of the modal coefficient is the best choice to be the label, and MSE or RMS will simply be the loss to train and predict the model. 
Most of the MD studies used a small number of test sets. Samples for comparisons on coefficients and images are selectively chosen, without any standards or metrics. In this paper, 10,000 test samples evaluated the performance of the trained model, and quartiles based on loss have been selected as an example to be shown. 
As the labels are not discrete but continuous, numerous samples are required to exactly predict the MD system. For example, if each label is divided into five levels, - which leads to under ~20\% of loss - samples of $5^n$ will be required, where $n$ is the number of labels. Four modes need seven labels, which requires 78125 samples. However, as the mode theory is studied intensively, it is available to generate training sets. If the number of modes increases, the required training set for model prediction will exponentially increase. Assuming one linear polarization and using $LP$ series will reduce the label by half. However, the $LP$ based model will fail to examine the ratio of cylindrical vector modes such as $TM_{01}$ mode or $TE_{01}$ mode.
In addition, if one desires to use two-dimensional input image, connecting multiple images horizontally (or vertically) - which generates 61$\times$(121*n)$\times$1 image in the case of the paper - will perform similar results. The depth of the CNN layers must be reduced, as the 3D CNN layer calculates the third dimension of the tensors at once.

\section{Conclusion}
The paper has proposed CNN model for MD on degenerated modes based on four or seven sets of input images from LPs and CPs. The CPs detect the direction of circularly polarized light and enhances the performance of the CNN model. $LP_{11}$ series has been decomposed from 7 images including the full image and images from four LPs of 0$^\circ$, 45$^\circ$, 90$^\circ$, and 135$^\circ$ and two CPs, RHCP and LHCP. Output labels of the model have been chosen to be the real and imaginary parts of mode coefficients. The model was trained via 100,000 training sets, and 10,000 test sets were evaluated. The trained model showed 0.0634 of label RMS, 0.0292 of intensity RMS, 0.1867 rad of phase MAE, and 0.9978 of average field correlation for case of 4 image sets. The performance of 4 image sets showed at least 50.68\% of performance enhancement compared to models considering only images after LPs. Quartiles in terms of label RMS have been selected to visualize the performance. The predicted values showed relatively reasonable match compared to the actual value.
The study is believed to be applied to obtain the coefficient of cylindrical vector mode. In addition, the proposed methods to decompose degenerate modes, select the loss function, and standards to select examples are expected to assist further CNN based MD studies or related works. 

\bibliography{sample}

\end{document}